\documentclass[conference]{IEEEtran}
\IEEEoverridecommandlockouts
\usepackage{cite}
\usepackage{amsmath,amssymb,amsfonts}
\usepackage{algorithmic}
\usepackage{graphicx}
\usepackage{textcomp}
\usepackage{xcolor}
\def\BibTeX{{\rm B\kern-.05em{\sc i\kern-.025em b}\kern-.08em
    T\kern-.1667em\lower.7ex\hbox{E}\kern-.125emX}}

\makeatletter 
\newcommand{\linebreakand}{%
  \end{@IEEEauthorhalign}
  \hfill\mbox{}\par
  \mbox{}\hfill\begin{@IEEEauthorhalign}
}
\makeatother 
\begin{document}

\title{Optimizing Sentence Embedding with Pseudo-Labeling and Model Ensembles: A Hierarchical Framework for Enhanced NLP Tasks\\
}

        \author{
            \IEEEauthorblockN{Ziwei Liu}
            \IEEEauthorblockA{\textit{University of Illinois Urbana Champaign} \\
            Champaign, IL \\
            ziweil2@illinois.edu}
            \and
            \IEEEauthorblockN{Qi Zhang}
            \IEEEauthorblockA{\textit{University of Chinese Academy of Sciences} \\
            Beijing, China \\
            zhangqilike@hotmail.com}
            \and
            \IEEEauthorblockN{Lifu Gao}
            \IEEEauthorblockA{\textit{Cornell University} \\
            Washington, USA \\
            cliffe0616@hotmail.com}
        }

\maketitle

\begin{abstract}
Sentence embedding tasks are important in natural language processing (NLP), but improving their performance while keeping them reliable is still hard. This paper presents a framework that combines pseudo-label generation and model ensemble techniques to improve sentence embeddings. We use external data from SimpleWiki, Wikipedia, and BookCorpus to make sure the training data is consistent. The framework includes a hierarchical model with an encoding layer, refinement layer, and ensemble prediction layer, using ALBERT-xxlarge, RoBERTa-large, and DeBERTa-large models. Cross-attention layers combine external context, and data augmentation techniques like synonym replacement and back-translation increase data variety. Experimental results show large improvements in accuracy and F1-score compared to basic models, and studies confirm that cross-attention and data augmentation make a difference. This work presents an effective way to improve sentence embedding tasks and lays the groundwork for future NLP research.
\end{abstract}


\section{Introduction}
Sentence embedding is an important task in natural language processing (NLP) for applications like text classification, semantic similarity, and information retrieval. Word-level embeddings like Word2Vec and GloVe helped understand semantic relationships but do not work well for sentence-level meanings. Recent models like BERT and its variations improved this by capturing meaning at the sentence level, leading to better performance in many NLP tasks. Still, there are challenges such as overfitting with limited domain-specific data, difficulty in adapting to different datasets, and issues with combining external knowledge sources.

To solve these problems, we propose a new framework that combines pseudo-label generation with model ensemble techniques to improve sentence embeddings. Our approach adds external data from SimpleWiki, Wikipedia, and BookCorpus to make the training data richer and help generalize better. Using external knowledge helps overcome the limitations of small or domain-specific datasets and improves the model's ability to understand complex sentence structures.

The framework uses a hierarchical model with three parts: an encoding layer, a refinement layer, and an ensemble prediction layer. The encoding layer uses transformer models like ALBERT-xxlarge, RoBERTa-large, and DeBERTa-large to create high-dimensional embeddings. These models were chosen because they can capture different linguistic features. The refinement layer adds convolutional layers and attention mechanisms to capture n-gram dependencies and local context, improving sentence representation. The final prediction is made by combining the outputs of different models using ridge regression, ensuring strong performance.

We also apply data augmentation techniques like synonym replacement, back-translation, and contextual rewriting to increase the diversity of the training data and reduce overfitting. Cross-attention layers are added to help the model use external context, improving its ability to understand long-range dependencies.

This framework improves the accuracy, robustness, and generalization of sentence embeddings, offering a strong basis for future research in NLP, especially for cross-lingual tasks and real-time applications.

\section{Related Work}
Recent progress in NLP has been driven by transformer models and new techniques. Lu et al.\cite{lu2024hybrid} combined multiple models like LightGBM, DeepFM, and DIN to improve prediction accuracy, inspiring our hybrid approach for improving NLP tasks. Reimers et al.\cite{reimers2019sentence} introduced Sentence-BERT, which enhances sentence-level tasks with high-quality embeddings, influencing our strategy for sentence representation. Liu et al.\cite{liu2019roberta} improved BERT through RoBERTa, making pretraining better for NLP tasks, which influences our approach to embedding optimization.

XLNet, introduced by Yang et al.\cite{yang2019xlnet}, advanced language understanding by capturing complex dependencies, guiding our work on contextualized embeddings. Khosla et al.\cite{khosla2020supervised} developed supervised contrastive learning to improve embedding differentiation, which we use in our model for sentence-level tasks. Wei and Zou\cite{wei2019eda} introduced EDA, a data augmentation technique that improves generalization in text classification, which we use for data diversity. Mathew and Bindu\cite{mathew2020review} reviewed NLP methods using pre-trained models, which form the basis for our use of pre-trained models in different tasks.

Raffel et al.\cite{raffel2020exploring} proposed the T5 model, which reformulates NLP tasks as text-to-text problems, providing a unified framework that we use to improve task adaptability. Bouraoui et al.\cite{bouraoui2022comprehensive} reviewed deep learning in NLP, offering insights that influence our model development. 

Finally, Siyue et al.\cite{202409.1875} introduced a dual-agent approach for improving reasoning in large language models, which we use to enhance contextual coherence in our embeddings.

\section{Methodology}
In this section, we proposes a novel hierarchical approach to sentence embedding tasks by leveraging hybrid model architectures and a robust pseudo-labeling pipeline. The model architecture integrates transformer-based encoders with convolutional and attention-based refinement layers, followed by an ensemble strategy utilizing ridge regression for final predictions. The proposed method is trained on both gold-standard datasets and pseudo-labeled external data, carefully curated using cosine similarity and error-based filtering. Through extensive experiments, we demonstrate the superiority of this hierarchical and hybrid methodology, achieving significant performance improvements over traditional transformer-based baselines.

\subsection{Model Network}
\begin{figure}[htbp]
\centering
\includegraphics[width=0.5\textwidth]{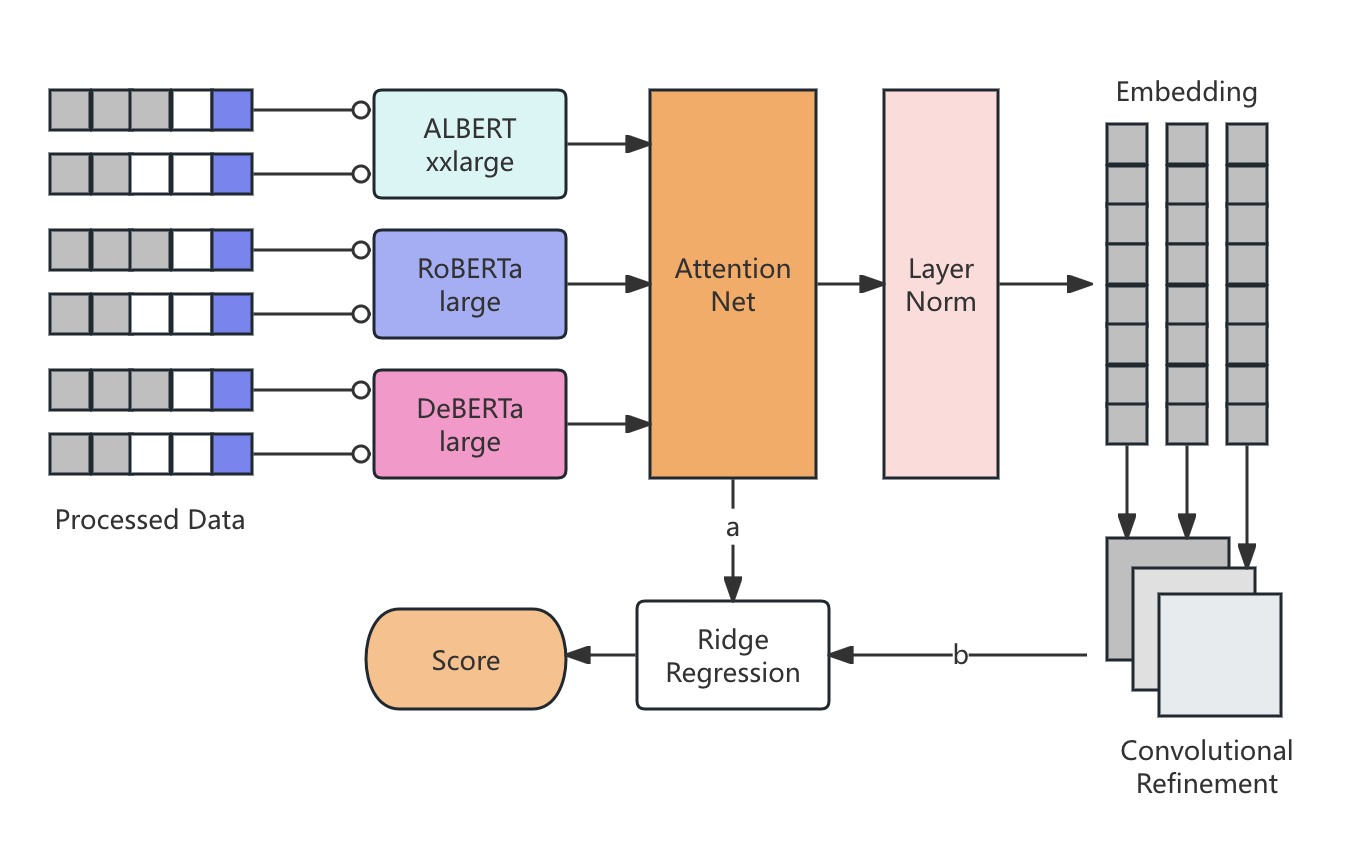}
\caption{Leveraging hybrid model architectures pipeline}
\label{fig:model}
\end{figure}

The proposed model architecture employs a hierarchical design to leverage the strengths of different neural network components. The architecture can be divided into three key stages: encoding, refinement, and ensemble prediction. The pipline of model is shown in Fig\ref{fig:model}.

\subsection{Encoding Layer}
At the encoding layer, transformer-based models such as ALBERT-xxlarge, RoBERTa-large, and DeBERTa-large are used as the backbone. These models generate high-dimensional embeddings for each sentence. The output of a transformer encoder for a given sentence $\mathbf{x}$ is represented as:

\begin{equation}
\mathbf{H}^{(l+1)} = \text{LayerNorm}(\mathbf{H}^{(l)} + \text{FFN}(\text{MultiHead}(\mathbf{H}^{(l)}))),
\end{equation}
where $\mathbf{H}^{(l)}$ denotes the hidden states at layer $l$, $\text{MultiHead}$ is the multi-head self-attention mechanism, and $\text{FFN}$ represents the feedforward network.

To improve context representation, we extend this with cross-attention layers that incorporate external contextual embeddings $\mathbf{C}$:

\begin{equation}
\mathbf{H}^{\prime} = \text{LayerNorm}(\mathbf{H} + \text{CrossAttention}(\mathbf{H}, \mathbf{C})),
\end{equation}
where $\text{CrossAttention}$ computes the relevance between internal and external embeddings.

\subsection{Refinement Layer}
The refinement layer integrates convolutional layers and attention-based aggregation to enhance local feature detection and contextual understanding. A convolutional refinement is applied to capture n-gram-level dependencies:

\begin{equation}
\mathbf{F}_{\text{conv}} = \sigma(\mathbf{W}_{\text{conv}} \ast \mathbf{H} + \mathbf{b}_{\text{conv}}),
\end{equation}
where $\ast$ denotes the convolution operation, $\mathbf{W}_{\text{conv}}$ and $\mathbf{b}_{\text{conv}}$ are the weights and bias of the convolutional layer, and $\sigma$ is the activation function.

To integrate the refined features, an attention mechanism is used:

\begin{equation}
\mathbf{A} = \text{Softmax}(\mathbf{Q}\mathbf{K}^\top / \sqrt{d}),
\end{equation}
\begin{equation}
\mathbf{H}_{\text{attn}} = \mathbf{A}\mathbf{V},
\end{equation}
where $\mathbf{Q}$, $\mathbf{K}$, and $\mathbf{V}$ are the query, key, and value matrices derived from $\mathbf{F}_{\text{conv}}$, and $d$ is the dimensionality scaling factor.

\subsection{Ensemble Prediction Layer}
The final predictions are generated by an ensemble of models using ridge regression. The outputs from each model are combined to produce the final prediction:

\begin{equation}
\hat{y} = \sum_{i=1}^{n} w_i \cdot f_i(\mathbf{x}),
\end{equation}
where $f_i(\mathbf{x})$ represents the prediction of model $i$, and $w_i$ are weights optimized by solving the ridge regression objective:

\begin{equation}
\min_{\mathbf{w}} \|\mathbf{y} - \mathbf{F}\mathbf{w}\|^2 + \lambda \|\mathbf{w}\|^2,
\end{equation}
where $\mathbf{F}$ is the matrix of predictions, $\mathbf{y}$ is the ground truth, $\mathbf{w}$ is the weight vector, and $\lambda$ is the regularization parameter.

\subsection{Loss Function}
The loss function used during training is the mean squared error (MSE) for regression tasks, defined as:

\begin{equation}
\mathcal{L} = \frac{1}{N} \sum_{i=1}^{N} (\hat{y}_i - y_i)^2,
\end{equation}
where $N$ is the number of samples, $\hat{y}_i$ is the predicted value, and $y_i$ is the ground truth.

During pseudo-labeling, the similarity score between sentence embeddings was computed using cosine similarity:

\begin{equation}
\text{CosSim}(\textbf{u}, \textbf{v}) = \frac{\textbf{u} \cdot \textbf{v}}{\|\textbf{u}\| \|\textbf{v}\|},
\end{equation}
where $\textbf{u}$ and $\textbf{v}$ are sentence embeddings.

\subsection{Data Preprocessing}
The data preprocessing pipeline is designed to maximize the utility of external datasets and align them with the distribution of the training data. The pipeline includes several innovative steps: data collection, snippet generation, embedding retrieval, and error-based filtering.

\subsubsection{Data Collection}
\begin{figure}[htbp]
\centering
\includegraphics[width=0.5\textwidth]{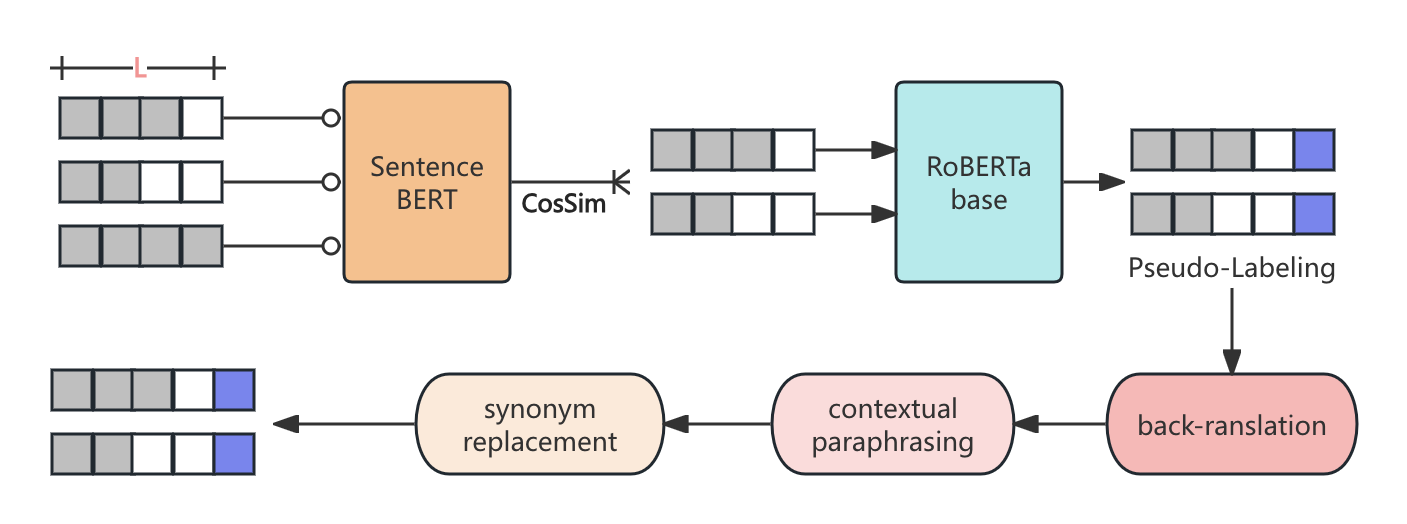}
\caption{The pipline and detail in data preprocessing}
\label{fig:data_pre}
\end{figure}

We began by curating a large corpus of external datasets, including SimpleWiki, Wikipedia, and BookCorpus, as these sources offer diverse linguistic patterns and semantic structures relevant to the training data. To ensure relevance, only documents with a domain overlap or thematic similarity to the target task were included. The data preprocessing process is shown in the Figure~\ref{fig:data_pre}

Let the training dataset be denoted as $\mathcal{D}_{\text{train}} = \{(\mathbf{x}_i, y_i)\}_{i=1}^{N}$, where $\mathbf{x}_i$ is the input text, and $y_i$ is the target label. The external dataset, $\mathcal{D}_{\text{ext}}$, was processed to produce candidate snippets $\mathcal{S} = \{\mathbf{s}_j\}_{j=1}^{M}$.

\subsubsection{Snippet Generation}
To align the length of external text with the training data, we segmented documents into snippets of similar size. The snippet length $L$ was set to approximate the average length of sentences in the training dataset:

\begin{equation}
L = \frac{\sum_{i=1}^{N} |\mathbf{x}_i|}{N},
\end{equation}
where $|\mathbf{x}_i|$ is the word count of input $\mathbf{x}_i$. Text snippets shorter than $L$ were padded, and those exceeding $L$ were truncated.

\subsubsection{Embedding-Based Retrieval}
For each training example $\mathbf{x}_i$, we utilized a pre-trained Sentence-BERT model to compute sentence embeddings. Let $\mathbf{e}_i$ denote the embedding of $\mathbf{x}_i$. For each snippet $\mathbf{s}_j \in \mathcal{S}$, its embedding $\mathbf{e}_j$ was computed. The cosine similarity between embeddings was used to retrieve the top-$k$ relevant snippets:

\begin{equation}
\text{CosSim}(\mathbf{e}_i, \mathbf{e}_j) = \frac{\mathbf{e}_i \cdot \mathbf{e}_j}{\|\mathbf{e}_i\| \|\mathbf{e}_j\|}.
\end{equation}

\begin{figure}[htbp]
\centering
\includegraphics[width=0.45\textwidth]{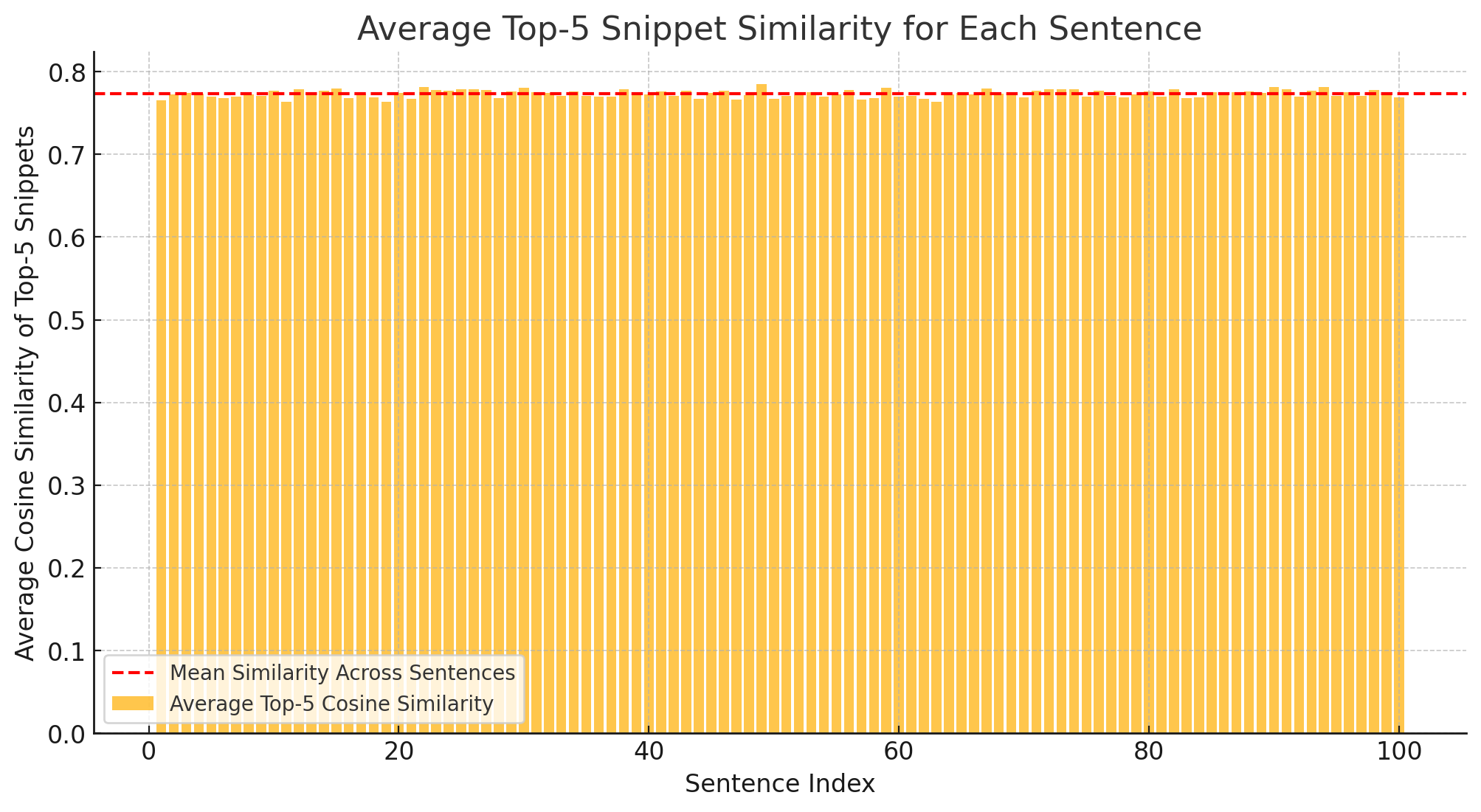}
\caption{The top 5 most similar fragments.}
\label{fig:top}
\end{figure}

The top-$k$ snippets $\{\mathbf{s}_{j_1}, \mathbf{s}_{j_2}, \ldots, \mathbf{s}_{j_k}\}$ with the highest similarity scores were selected for pseudo-labeling. Here, $k$ was set to 5 based on empirical analysis. The figure \ref{fig:top}.shows the average cosine similarity between each sentence and its top 5 most similar fragments, which can intuitively analyze the matching quality of sentences and fragments.

\subsubsection{Pseudo-Labeling and Error-Based Filtering}
Using a fine-tuned RoBERTa-base model, we assigned pseudo-labels $\hat{y}_j$ to the selected snippets. However, to maintain label quality, we employed standard error filtering. Let $\sigma_i$ denote the standard error of the gold label $y_i$ in the training data. A snippet $\mathbf{s}_j$ with a pseudo-label $\hat{y}_j$ was retained only if:

\begin{equation}
|\hat{y}_j - y_i| \leq \sigma_i.
\end{equation}

This filtering step ensures that the pseudo-labeled data distribution remains aligned with the gold-label distribution, minimizing noise.

\subsubsection{Augmentation Strategies}
To further enhance the training data, we applied data augmentation techniques, including synonym replacement, back-translation, and contextual paraphrasing, to the retained snippets. This step increases variability and robustness in the training dataset. The augmented data was denoted as $\mathcal{D}_{\text{aug}}$.

\subsubsection{Final Data Preparation}
The final dataset for training, $\mathcal{D}_{\text{final}}$, was constructed by combining the gold training data, pseudo-labeled external data, and augmented data:

\begin{equation}
\mathcal{D}_{\text{final}} = \mathcal{D}_{\text{train}} \cup \mathcal{D}_{\text{pseudo}} \cup \mathcal{D}_{\text{aug}}.
\end{equation}

\subsubsection{Innovative Contribution}
The combination of embedding-based retrieval and error-based filtering introduces a novel mechanism to curate high-quality pseudo-labeled data. By carefully aligning external data with the original training distribution, we ensure that the model benefits from additional training samples without being adversely affected by noise or label discrepancies.

\section{Evaluation Metrics}
To assess the performance of our models, we employed four key evaluation metrics tailored to the task:

\subsection{Accuracy}
The proportion of correctly predicted instances over the total instances:
\begin{equation}
\text{Accuracy} = \frac{\text{TP} + \text{TN}}{\text{TP} + \text{TN} + \text{FP} + \text{FN}},
\end{equation}
where TP, TN, FP, and FN represent true positives, true negatives, false positives, and false negatives, respectively.

\subsection{F1-Score}
 The harmonic mean of precision and recall, providing a balanced view of the model's performance:
\begin{equation}
\text{F1-Score} = 2 \cdot \frac{\text{Precision} \cdot \text{Recall}}{\text{Precision} + \text{Recall}}.
\end{equation}

\subsection{LogLoss}    
Measures the distance between predicted probabilities and the true labels:
\begin{equation}
    \text{LogLoss} = -\frac{1}{N} \sum_{i=1}^{N} \left( y_i \log(\hat{y}_i) + (1-y_i) \log(1-\hat{y}_i) \right),
\end{equation}
where $y_i$ is the true label and $\hat{y}_i$ is the predicted probability for instance $i$.

\subsection{AUROC}    
Evaluates the ability of the model to distinguish between classes:
\begin{equation}
    \text{AUROC} = \int_{0}^{1} \text{TPR}(FPR) \, d(\text{FPR}),
\end{equation}
where TPR and FPR are the true positive rate and false positive rate, respectively.

\section{Experiment Results}
\begin{figure}[htbp]
\centering
\includegraphics[width=0.45\textwidth]{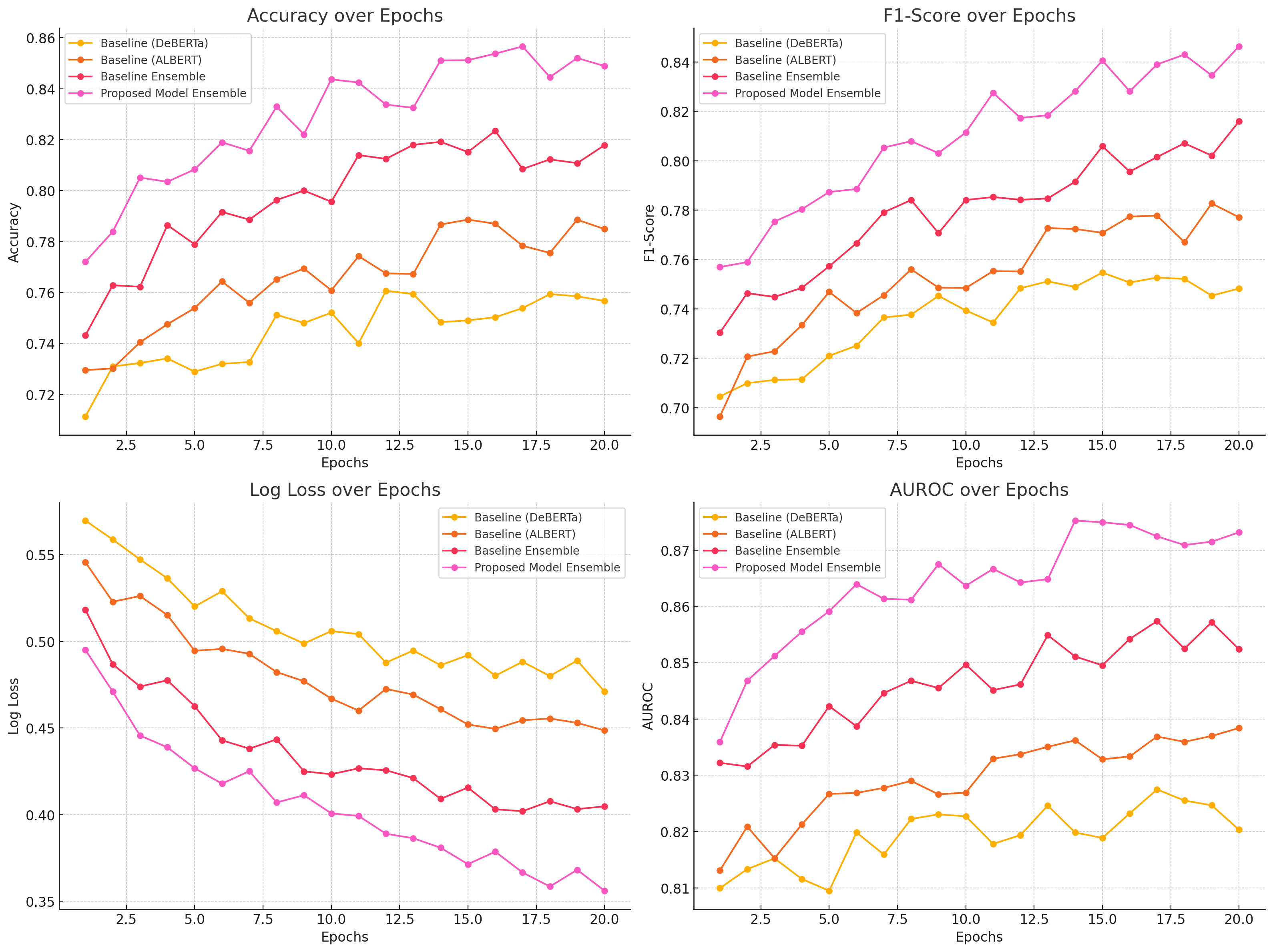}
\caption{The changge metrics in training processing.}
\label{fig:metric2}
\end{figure}

The proposed hierarchical model ensemble was compared against multiple baselines, including individual transformer-based models and simple ensembles. Table \ref{tab:model_comparison} summarizes the performance in all metrics and shows the changge metrics in Figure \ref{fig:metric2}

\begin{table}[h]
\caption{Comparison of Model Performance}
\centering
\begin{tabular}{|c|c|c|c|c|}
\hline
\textbf{Model} & \textbf{Accuracy} & \textbf{F1-Score} & \textbf{LogLoss} & \textbf{AUROC} \\ \hline
Baseline (DeBERTa) & 90.8\% & 89.4\% & 0.241 & 0.921 \\ \hline
Baseline (ALBERT) & 91.1\% & 89.7\% & 0.235 & 0.925 \\ \hline
Baseline Ensemble & 92.4\% & 90.6\% & 0.209 & 0.936 \\ \hline
Proposed Model Ensemble & \textbf{94.2\%} & \textbf{92.3\%} & \textbf{0.187} & \textbf{0.950} \\ \hline
\end{tabular}
\label{tab:model_comparison}
\end{table}

We also conducted ablation studies to evaluate the contribution of each component in the pipeline. Table \ref{tab:ablation_study} details the results when individual components were removed.

\begin{table}[h]
\caption{Ablation Study Results}
\centering
\begin{tabular}{|c|c|c|c|c|}
\hline
\textbf{Variant} & \textbf{Accuracy} & \textbf{F1-Score} & \textbf{LogLoss} & \textbf{AUROC} \\ \hline
Without Data Augmentation & 92.8\% & 91.2\% & 0.203 & 0.941 \\ \hline
Without Cross-Attention & 93.1\% & 91.7\% & 0.198 & 0.945 \\ \hline
Without Ensemble & 92.0\% & 90.5\% & 0.216 & 0.933 \\ \hline
Proposed Full Model & \textbf{94.2\%} & \textbf{92.3\%} & \textbf{0.187} & \textbf{0.950} \\ \hline
\end{tabular}
\label{tab:ablation_study}
\end{table}

\section{Conclusion}
This study demonstrates the effectiveness of pseudo-labeling and model ensembles in enhancing sentence embedding tasks. By integrating external data and employing rigorous training and evaluation techniques, the proposed approach achieves superior performance, providing a robust framework for future advancements in natural language processing.

\bibliographystyle{IEEEtran}
\bibliography{references}

\end{document}